\newif\ifanon
\anonfalse

\documentclass[letterpaper]{article} 
\ifanon
    \usepackage[submission]{aaai25}  
\else
    \usepackage{aaai25}  
\fi
\usepackage{times}  
\usepackage{helvet}  
\usepackage{courier}  
\usepackage[hyphens]{url}  
\usepackage{graphicx} 
\usepackage{natbib}  
\usepackage{caption} 
\usepackage{algorithm}
\usepackage{algorithmic}

\usepackage{newfloat}
\usepackage{listings}
\DeclareCaptionStyle{ruled}{labelfont=normalfont,labelsep=colon,strut=off} 
\floatstyle{ruled}
\newfloat{listing}{tb}{lst}{}
\floatname{listing}{Listing}
\usepackage{xcolor}
\usepackage{booktabs}
\usepackage{subcaption}
\usepackage{amsmath}
\usepackage{mathrsfs}
\usepackage{xparse}
\usepackage{etoolbox}     
\usepackage{pgffor} 
\usepackage{xspace}
\usepackage{amssymb}

\ifanon
  \newcommand{\github}{https://github.com\xspace}
\else
  \newcommand{\github}{https://github.com/RedForestAi/WebEyeTrack\xspace} 
\fi

\title{\webeyetrack: Scalable Eye-Tracking for the Browser via On-Device Few-Shot Personalization}
\author{
    Eduardo Davalos\textsuperscript{\rm 1}\equalcontrib,
    Yike Zhang\textsuperscript{\rm 2}\equalcontrib,
    Namrata Srivastava\textsuperscript{\rm 3},
    Yashvitha Thatigotla\textsuperscript{\rm 3},
    Jorge A. Salas\textsuperscript{\rm 3},
    Sara McFadden\textsuperscript{\rm 3},
    Sun-Joo Cho\textsuperscript{\rm 3},
    Amanda Goodwin\textsuperscript{\rm 3},
    Ashwin TS\textsuperscript{\rm 3},
    Gautam Biswas\textsuperscript{\rm 3}
}
\affiliations{
    \textsuperscript{\rm 1}Trinity University, San Antonio, TX, USA\\
    \textsuperscript{\rm 2}St. Mary's University, San Antonio, TX, USA\\
    \textsuperscript{\rm 3}Vanderbilt University, Nashville, TN, USA\\
    davalosedu515@trinity.edu, yzhang5@stmarytx.edu, namrata.srivastava@vanderbilt.edu, yashvitha.thatigotla@vanderbilt.edu, jorge.a.salas@vanderbilt.edu, sara.mcfadden@vanderbilt.edu, sj.cho@vanderbilt.edu, ashwindixit9@gmail.com, amanda.goodwin@vanderbilt.edu, gautam.biswas@vanderbilt.edu
}

\usepackage{bibentry}

\newcommand{\webeyetrack}{\textsc{WebEyeTrack}\xspace}

\begin{document}


\maketitle

\begin{abstract}

With advancements in AI, new gaze estimation methods are exceeding state-of-the-art (SOTA) benchmarks, but their real-world application reveals a gap with commercial eye-tracking solutions. Factors like model size, inference time, and privacy often go unaddressed. Meanwhile, webcam-based eye-tracking methods lack sufficient accuracy, in particular due to head movement. To tackle these issues, we introduce \textbf{WebEyeTrack}, a framework that integrates lightweight SOTA gaze estimation models directly in the browser. It incorporates model-based head pose estimation and on-device few-shot learning with as few as nine calibration samples ($k \leq 9$). WebEyeTrack adapts to new users, achieving SOTA performance with an error margin of $2.32$ cm on GazeCapture and real-time inference speeds of $2.4$ milliseconds on an iPhone 14. Our open-source code is available at \github.
\end{abstract}

%

\begin{links}
    \ifanon
      \link{Code}{https://github.com} (anonymized)
    \else
      \link{Code}{https://github.com/RedForestAI/WebEyeTrack}
    \fi
    \ifanon
      \link{Demo}{https://azure-olympie-5.tiiny.site} (anonymized)
    \else
    \fi
    \ifanon
      \link{Website}{https://site.github.io} (anonymized)
    \else
      \link{Website}{https://redforestai.github.io/WebEyeTrack/}
    \fi
\end{links}

\section{Introduction}

Eye-tracking has been a transformative tool for investigating human-computer interactions, as it uncovers subtle shifts in visual attention \cite{hyona_commentary_2003}. However, its reliance on expensive specialized hardware, such as EyeLink 1000 and Tobii Pro Fusion has confined most gaze-tracking research to controlled laboratory environments \cite{heck_webcam_2023}. Similarly, virtual reality solutions like the Apple Vision Pro remain financially out of reach for widespread use. These limitations have hindered the scalability and practical application of gaze-enhanced technologies and feedback systems.

\begin{figure}
    \centering
    \includegraphics[clip, trim={0.5cm 0 0.5cm 0}, width=\linewidth]{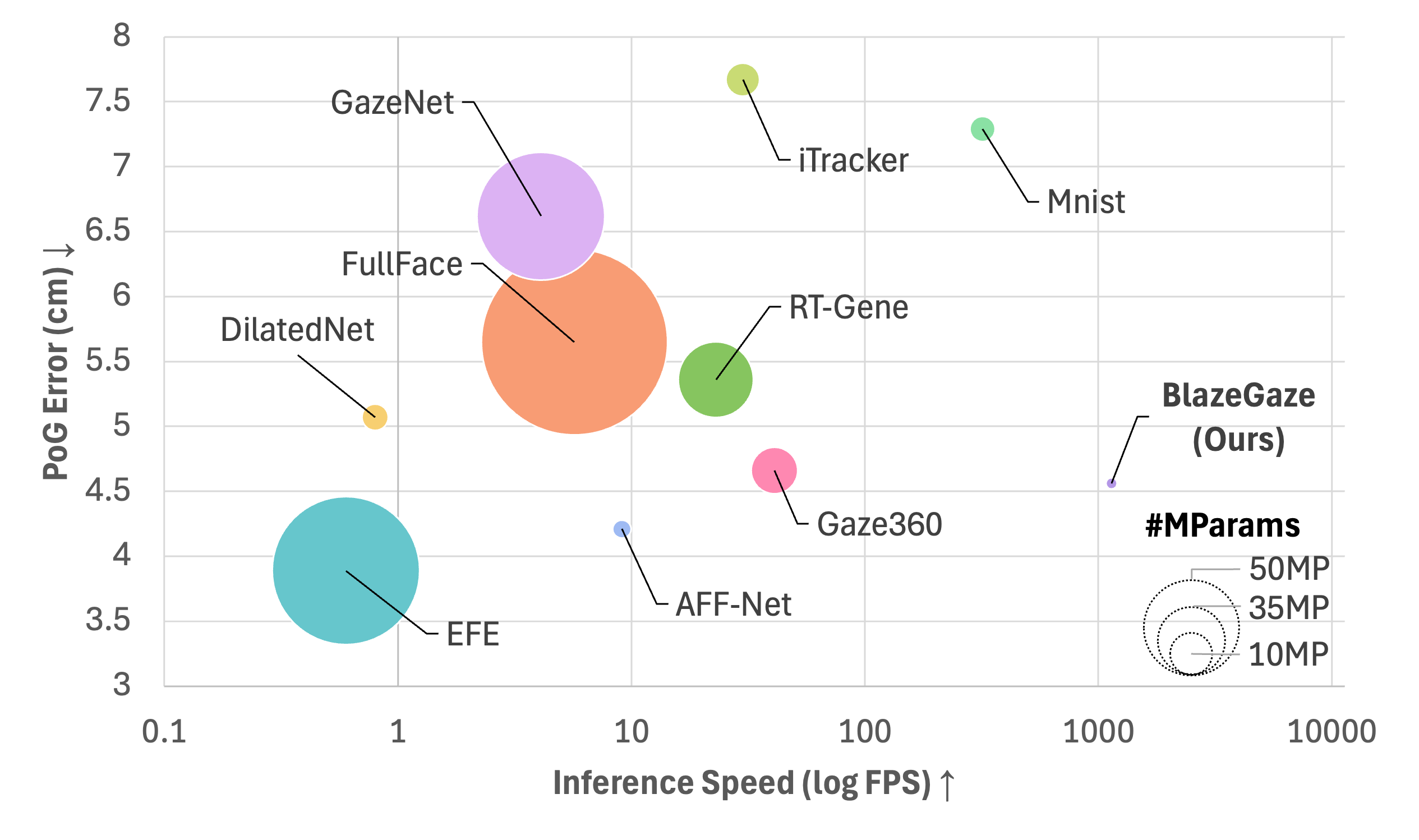}
    \caption{\textbf{Point-of-Gaze (PoG) Error (cm) vs. Inference Speed (log-scale FPS) Across Gaze Estimation Methods}: BlazeGaze (ours) achieves high accuracy with orders of magnitude faster inference speed using lightweight CNN BlazeBlocks.}
    \label{fig:performance_vs_inference_graph}
\end{figure}

To reduce reliance on specialized hardware, researchers have actively pursued webcam-based eye-tracking solutions that utilize built-in cameras on consumer devices. Two key areas of focus in this field are appearance-based gaze estimation and webcam-based eye-tracking, both of which have made significant advancements using standard monocular cameras \cite{cheng_appearance-based_2021}. For example, recent appearance-based methods have shown improved accuracy on commonly used gaze estimation datasets such as MPIIGaze \cite{zhang_appearance-based_2015}, MPIIFaceGaze \cite{zhang_its_2016}, and EyeDiap \cite{alberto_funes_mora_eyediap_2014}. However, many of these AI models primarily aim to achieve state-of-the-art (SOTA) performance without considering practical deployment constraints. These constraints include varying display sizes, computational efficiency, model size, ease of calibration, and the ability to generalize to new users. While some efforts have successfully integrated gaze estimation models into comprehensive eye-tracking solutions \cite{heck_webcam_2023}, achieving real-time, fully functional eye-tracking systems remains a considerable technical challenge. Retrofitting existing models that do not address these design considerations often involves extensive optimization and may still fail to meet practical requirements. As a result, state-of-the-art gaze estimation methods have not yet been broadly implemented, mainly due to the difficulties of running these AI models on resource-constrained devices.

At the same time, webcam-based eye-tracking methods have taken a practical approach, addressing real-world deployment challenges \cite{heck_webcam_2023}. These solutions are often tied to specific software ecosystems and toolkits, hindering portability to platforms such as mobile devices or web browsers. As web applications gain popularity for their scalability, ease of deployment, and cloud integration \cite{Shukla_2023}, tools like WebGazer \cite{papoutsaki_webgazer_2016} have emerged to support eye-tracking directly within the browser. However, many browser-friendly approaches rely on simple statistical or classical machine learning models \cite{heck_webcam_2023}, such as ridge regression \cite{xu_turkergaze_2015} or support vector regression \cite{papoutsaki_webgazer_2016}, and avoid 3D gaze reasoning to reduce computational load. While these techniques improve accessibility, they often compromise accuracy and robustness under natural head motion.

To bridge the gap between high-accuracy appearance-based gaze estimation methods and scalable webcam-based eye-tracking solutions, we introduce \webeyetrack, a few-shot, headpose-aware gaze estimation solution for the browser (Fig~\ref{fig:webeyetrack_framework}). \webeyetrack combines model-based headpose estimation (via 3D face reconstruction and radial procrustes analysis) with BlazeGaze, a lightweight CNN model optimized for real-time inference. It computes the Point-of-Gaze (PoG), detects eye state (open/closed) via the eye aspect ratio (EAR) \cite{soukupova_real-time_2016} to suppress predictions during blinks, implements continuous clickstream calibration, and provides online fixation detection via WebFixRT\footnote{https://github.com}, thereby creating a full-featured and web-friendly eye-tracking system. We provide both Python and client-side JavaScript implementations to support model development and seamless integration into research and deployment pipelines. In evaluations on standard gaze datasets, \webeyetrack achieves comparable SOTA performance and demonstrates real-time performance on mobile phones, tablets, and laptops. 

\noindent
Overall, our work makes the following contributions:
\begin{itemize}
    \item \webeyetrack: an open-source novel browser-friendly framework that performs few-shot gaze estimation with privacy-preserving on-device personalization and inference. 
    \item A novel model-based metric headpose estimation via face mesh reconstruction and radial procrustes analysis.
    \item BlazeGaze: A novel, 670KB CNN model based on BlazeBlocks that achieves real-time inference on mobile CPUs and GPUs.
\end{itemize}

\section{Background}

\subsubsection{Gaze Estimation}

Classical gaze estimation relied on model-based approaches for (1) 3D gaze estimation (predicting gaze direction as a unit vector), and (2) 2D gaze estimation (predicting gaze target on a screen). These methods used predefined eyeball models and extensive calibration procedures \cite{dongheng_li_starburst_2005,wood_eyetab_2014,brousseau_accurate_2018,wang_real_2017}. In contrast, modern appearance-based methods require minimal setup and leverage deep learning for improved robustness \cite{cheng_appearance-based_2021}.

The emergence of CNNs and datasets such as MPIIGaze \cite{zhang_appearance-based_2015}, GazeCapture \cite{krafka_eye_2016}, and EyeDiap \cite{alberto_funes_mora_eyediap_2014} has led to the development of 2D and 3D gaze estimation systems capable of achieving errors of 6–8 degrees and 3–7 centimeters \cite{zhang_appearance-based_2015}. Key techniques that have contributed to this progress include multimodal inputs \cite{krafka_eye_2016}, multitask learning \cite{leal-taixe_deep_2019}, self-supervised learning \cite{ferrari_appearance-based_2018}, data normalization \cite{zhang_revisiting_2018}, and domain adaptation \cite{li_evaluation_2020}. More recently, Vision Transformers have further enhanced accuracy, reducing error to 4.0 degrees and 3.6 centimeters \cite{cheng_gaze_2022}.
Despite strong within-dataset performance, generalization to unseen users remains poor \cite{cheng_appearance-based_2021}. Few-shot learning addresses this by adapting to new users with minimal labeled data. However, the few-shot FAZE framework \cite{park_few-shot_2019} requires substantial computational resources, including a powerful GPU for real-time inference, while SAGE \cite{he_-device_2019} is a closed-source implementation, restricting evaluation, usage, and practical adoption.

\subsubsection{Metric Head Pose Estimation}

Head pose is frequently utilized to improve gaze estimation \cite{zhang_its_2016}, but most methods rely on having access to ground-truth poses obtained from depth sensors or multi-camera setups \cite{alberto_funes_mora_eyediap_2014, sugano_learning-by-synthesis_2014}. These setups are often impractical for browser-based applications. Monocular methods tend to struggle with real-time performance and often produce scale-ambiguous poses \cite{kazemi_one_2014, guo_towards_2021, baltrusaitis_openface_2018}, which renders self-consistent head pose estimation unreliable.

\subsubsection{Webcam-based Eye-Tracking}

Deploying deep learning-based gaze systems in real-world applications is challenging due to high hardware requirements and compatibility issues with programming languages. Most implementations, such as OpenGaze \cite{zhang_evaluation_2019} and FAZE \cite{park_few-shot_2019}, are based on C++ and Python and require PyTorch and CUDA, which impose high computational demands unsuited to many consumer devices. Web-compatible tools like iTracker \cite{krafka_eye_2016} and iMon \cite{huynh_imon_2021} are often outdated or poorly documented. Additionally, using cloud-based training and inference raises concerns about privacy and latency \cite{park_gazel_2021}.

\subsubsection{Browser-based and Real-time Eye-Tracking}

Browser-compatible solutions such as RealEye \cite{pietrzak_realeye_2025}, WebET \cite{imotions_imotions_2023}, and GazeCloudAPI are closed-source and do not provide transparent benchmarking. Among open-source tools, WebGazer \cite{papoutsaki_webgazer_2016} is widely used, utilizing facial landmarks and ridge regression for estimating 2D PoG. However, its accuracy declines significantly over time due to its lack of head pose awareness, resulting in the error increasing from approximately 5 to 10 cm during a 20-minute test \cite{papoutsaki_eye_2018}.

We bridge this gap by integrating fast CNN-based gaze estimation, inspired by BlazeFace's BlazeBlocks \cite{bazarevsky_blazeface_2019}, with meta-learning for few-shot adaptation and real-time inference. Our model is also headpose-aware, improving robustness across users and environments.

\section{Method}

\begin{figure}
    \centering
    \includegraphics[width=\linewidth]{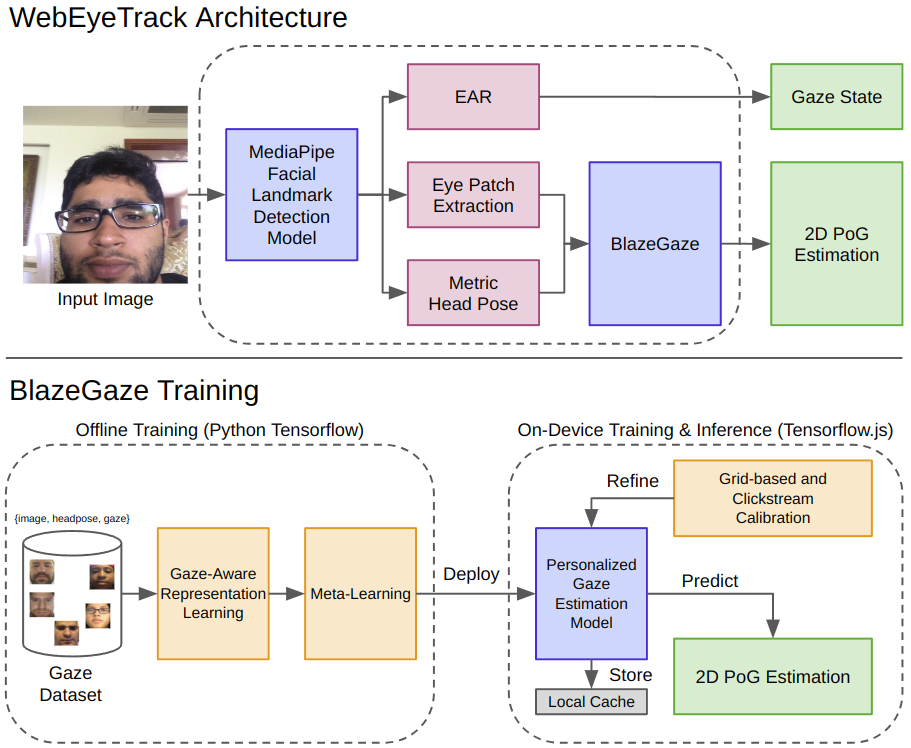}
    \caption{\textbf{\webeyetrack Framework Overview}: Framework composed of multiple model-based routines along with a CNN-based BlazeGaze gaze estimation model. The BlazeGaze model is trained to ensure privacy with on-device calibration and inference.}
    \label{fig:webeyetrack_framework}
\end{figure}

This section discusses \webeyetrack, a real-time gaze estimation framework designed to function effectively in challenging, real-world environments, such as web browsers and mobile devices. As shown in Fig. \ref{fig:webeyetrack_framework}, the computational pipeline consists of utilizing MediaPipe's facial landmark detection model to estimate the gaze state (open/close), eye patch region, and metric head pose. These variables are then passed to BlazeGaze, a lightweight 2D PoG estimation model. A two-step training process is used to derive an adaptive gaze meta-learner that is then deployed to user's browsers and perform on-device calibration, inference, and model caching. We aim to deliver personalized, head-pose-aware gaze predictions with low computational costs.

To accomplish this, \webeyetrack incorporates two main components: (1) a model-based 3D face reconstruction pipeline that accurately estimates head pose in metric space from monocular images, and (2) an appearance-based gaze estimation model called BlazeGaze, which utilizes model-agnostic meta-learning (MAML) to adapt to new users with minimal calibration data. This hybrid approach allows our system to use explicit geometric cues (such as head orientation and position) in conjunction with implicit appearance features (like eye-region encoding), resulting in improved generalization and personalization.

We first describe our novel metric 3D face reconstruction and headpose pipeline in Sec. \ref{sec:face_reconstruction}. This approach leverages MediaPipe's real-time Facial Landmark Detection model and an iris-based depth estimation algorithm to produce a fully metric head pose. We then detail the BlazeGaze model, which adopts a few-shot adaptation approach to learn a user-specific gaze regressor using only a handful of calibration samples ($k \leq 9$), in Sec. \ref{sec:blazegaze}. 

\subsection{Metric Face Reconstruction and Headpose}\label{sec:face_reconstruction}

3D face reconstruction is used to estimate head pose (position and orientation) in metric space rather than relative units. MediaPipe's Facial Landmark Detection model \cite{grishchenko_attention_2020} offers a lightweight pipeline for detecting 3D facial landmarks and reconstructing a canonical face mesh. The output UVZ landmark coordinates $\{x_i^I\}_{i=1}^N$ contain X and Y values normalized to screen space, while Z values are relative and scaled using a weak perspective projection. The model also provides a relative face transformation matrix \( P = [R \mid t] \), where $R \in \mathbb{R}^{3\times3}$ rotation matrix and $t\in \mathbb{R}^3$ translation vector transform the points from the canonical face coordinate space (FCS) to the camera coordinate space (CCS). The canonical mesh follows a fixed 468-point topology used for consistent face landmark estimation across frames.

In gaze-estimation datasets such as GazeCapture \cite{krafka_eye_2016}, the camera intrinsics are provided, which include the focal length $(f_x,f_y)$ along with the principal point $(c_x,c_y)$. This supports the construction of a camera intrinsics matrix:

\begin{equation}
    K = \begin{bmatrix}
        f_x & 0 & c_x \\
        0 & f_y & c_y \\
        0 & 0 & 1 
    \end{bmatrix}
\end{equation}

The first step is the conversion of facial landmarks from UVZ in image-plane space to XYZ space through reprojection, leveraging the camera intrinsics matrix:

\begin{equation}
    X = \frac{(u - c_x) \cdot z}{f_x}, \quad Y = \frac{(v - c_y) \cdot z}{f_y}
\end{equation}

To convert all 2D UVZ facial landmarks into 3D XYZ space, we reproject and normalize the points (centered around the nose [ID=4] and with face width set to 1 with leftmost [ID=356] and rightmost [ID=127] landmarks) using the following:

\begin{equation}
    \{x_i^R\}_{i=1}^N = \text{Reproject}(\{x_i^I\}_{i=1}^N, K^\prime)
\end{equation}

\begin{equation}
    x_i^F=\frac{x_i^R-x_{nose}^R}{\lVert x_{left}^R - x_{right}^R \rVert}
\end{equation}


\subsubsection{Face Scale Estimation}

The normalized XYZ facial landmarks \( \{x_i^F\}_{i=1}^N \) are unitless and relative, making them unsuitable for downstream tasks due to inconsistent depth scaling. To recover metric position and orientation (in centimeters), we first estimate face scale. Using a standard iris diameter of \( \alpha = 1.2 \) cm from gaze estimation literature \cite{wen_accurate_2020}, we compute face width from the pixel ratio between the iris diameter \( d_{\text{iris}}^I \) to face width \( d_w^I \). This ratio allows transforming the normalized points into metric landmarks \( \{ x_i^{F^\prime} \}_{i=1}^N \) via:

\begin{equation}
    \{x_i^{F^\prime}\}_{i=1}^N = \alpha \frac{d_w^I}{d_{iris}^I} \cdot \{ x_i^F \}_{i=1}^N
\end{equation}

\begin{figure}[t]
    \centering
    \includegraphics[width=\linewidth]{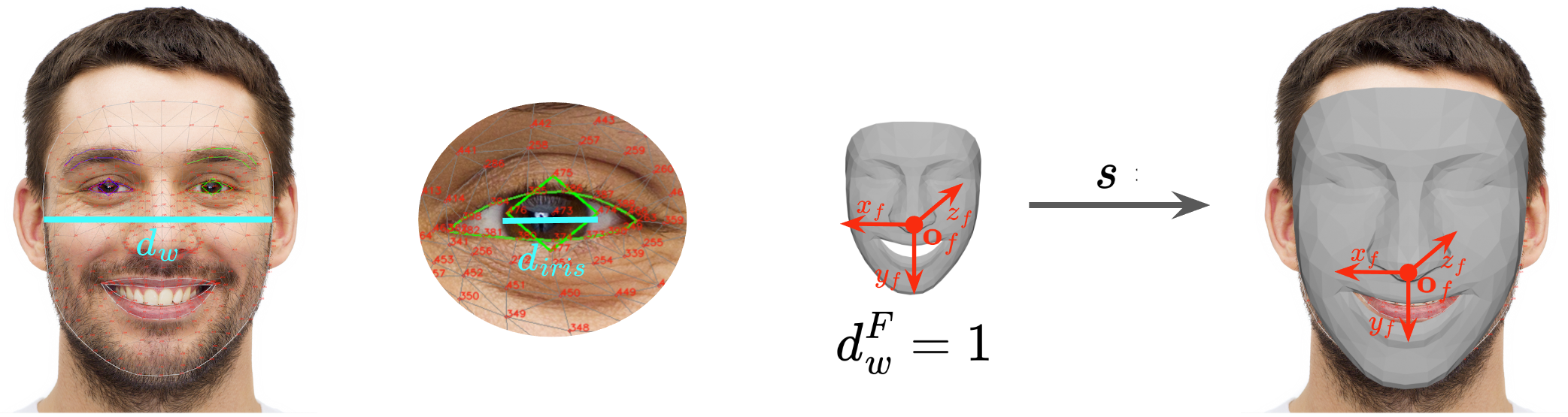}
    \caption{\textbf{Iris-based Face Scaling}: A constant iris diameter $\alpha$ enables metric face width estimation, which is applied to the normalized face mesh to approximate the user’s true facial dimensions.}
    \label{fig:enter-label}
\end{figure}

\subsubsection{Metric Headpose}

With metric facial landmarks, we estimate the 3D metric transformation matrix \( P = [R \mid t^\prime] \) representing head pose. The rotation matrix \( R \) from MediaPipe's relative pose output \( P = [R \mid t] \), is scale-invariant and can be reused to estimate the metric translation \( t^\prime \). We propose an iterative refinement algorithm based on our novel radial Procrustes alignment \cite{Gower_1975} and similar triangles. The goal is to minimize the Euclidean distance between the original UVZ landmarks and the newly projected 3D landmarks under the estimated transformation:

\begin{equation}
    t^\prime = \arg\min_{t^\prime} \sum_{i=1}^N \lVert x_i^I - K \cdot [R \mid t^\prime] \cdot x_i^{F^\prime}\rVert^2
\end{equation}


The algorithm initializes with depth \( z_0 = 60 \) cm, based on average monitor-to-face distance reported in \cite{cheng_appearance-based_2021}, and an initial XY estimate derived by reprojecting the nose’s UV position. This step aligns the UV center of projected and original UVZ points, reducing the problem to iterative depth refinement. A lightweight radial Procrustes method determines the direction and magnitude of the depth update by computing a unit direction vector \( v_i \) from the center \( c \) between matched projected points \( \hat{x}_i^I \) and original landmarks \( x_i^I \), using:

\begin{equation}
    v_i = \frac{x_i^I - \hat{x}_i^I}{\lVert x_i^I - \hat{x}_i^I \rVert}
\end{equation}

\begin{equation}
    z_{\text{update}} = \text{clip}\left(\frac{\beta}{N} \sum_{i=1}^N \text{sign}(v_i \cdot c_i) \cdot \lVert v_i \rVert, -\Delta_{\text{max}}, \Delta_{\text{max}} \right),
\end{equation} where $\beta$ is a scaling factor of $0.1$ and $\Delta_{\text{max}}$ is a threshold cap of 5 cm. A clip function acts as safety guardrails.

After computing and applying $z_{\text{update}}$, the corresponding metric XY values are updated via similar triangles to preserve the reprojection consistency, ensuring the projected nose center $\hat{x}_{\text{nose}}^I$ aligns with the observed nose center $\hat{x}_{\text{nose}}^I$. The iterative algorithm stops when \( z_{\text{update}} < 0.25\) cm or when the iteration cap of 10 is reached.

\subsection{BlazeGaze: Lightweight Personalized Gaze} \label{sec:blazegaze}

\begin{figure}
    \centering
    \includegraphics[width=\linewidth]{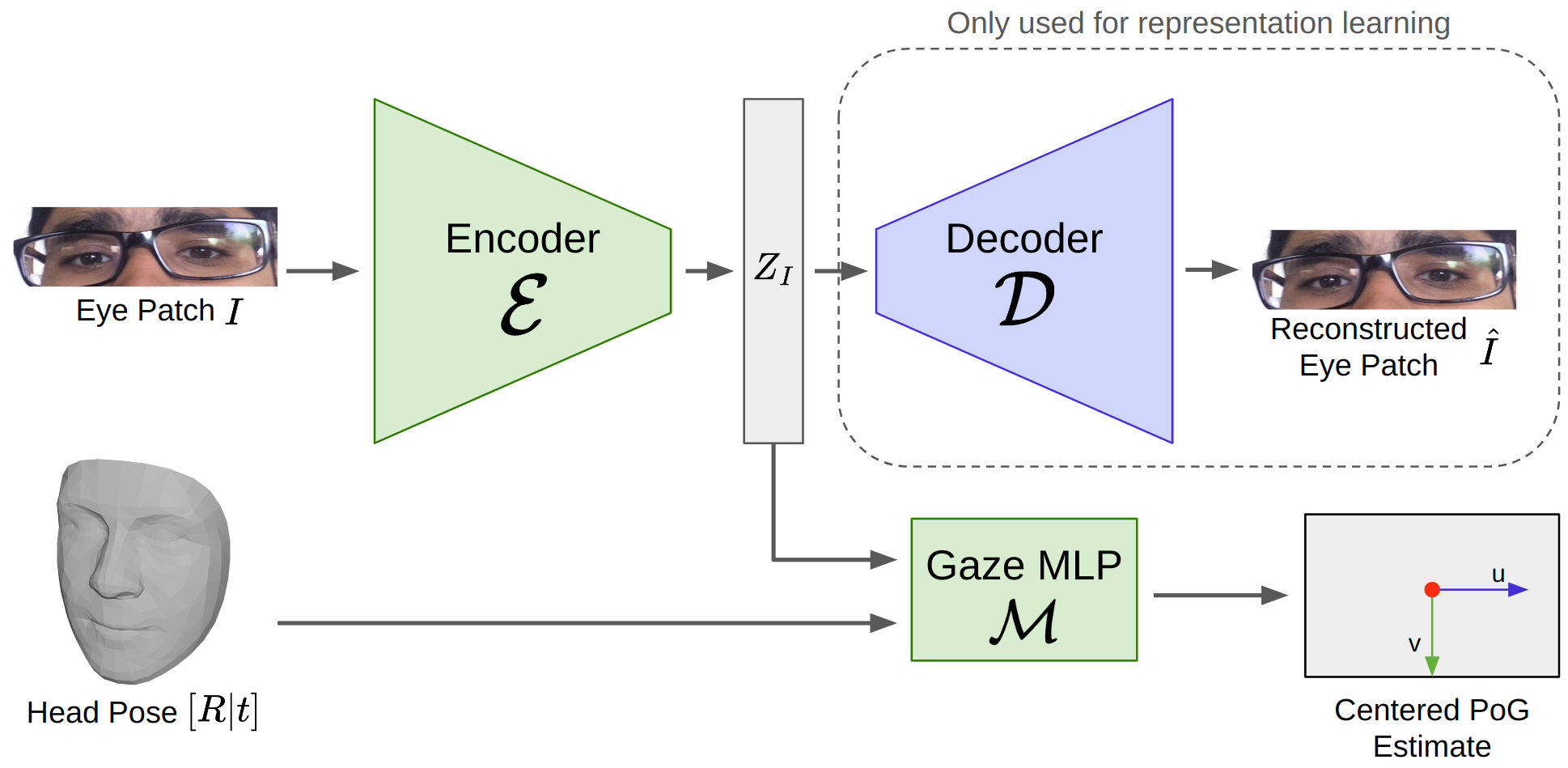}
    \caption{\textbf{BlazeGaze Architecture}: The BlazeGaze model takes as input the eye patch and the metric head pose and is composed of BlazeBlock-based encoder, decoder, and a gaze multi-layer perceptron (MLP). The decoder is used in representation learning, but is discarded during meta-learning.}
    \label{fig:blazegaze_architecture}
\end{figure}

As shown in Fig.~\ref{fig:blazegaze_architecture}, BlazeGaze consists of an encoder $\mathcal{E}:I\rightarrow\mathbf{z}$, decoder \( \mathcal{D}: \mathbf{z} \rightarrow \hat{I} \), and gaze estimator \( \mathcal{M}: (\mathbf{z}, [R \mid t]) \rightarrow \mathbf{g} \). We adopt a two-stage training strategy (see Fig.~\ref{fig:webeyetrack_framework}). In Stage 1, all components are trained jointly to learn a robust embedding via a standard train/validation split. In Stage 2, the decoder is discarded, the encoder is frozen, and the gaze estimator is adapted into a meta-learner using MAML to enable efficient user adaptation. This separation reduces the number of trainable parameters during meta-learning and avoids instability from joint optimization. To enhance cross-device generalization (e.g., mobile in GazeCapture vs. laptops in MPIIFaceGaze), gaze predictions are normalized to \( \mathbf{g} \in [[-0.5, 0.5]]^2 \) with the origin at screen center, enabling device-agnostic evaluation across varying screen resolutions and sizes.

\subsubsection{Gaze-Aware Representation Learning}

The first-stage of training uses a multi-objective loss function defined as:

\begin{equation}
    \mathcal{L}_{\text{all}} = \beta_{\text{r}} \mathcal{L}_{\text{r}} + \beta_{\text{g}} \mathcal{L}_{\text{g}} + \beta_{\text{c}} \mathcal{L}_{\text{c}},
\end{equation} %
where $\beta=\{\beta_{r}, \beta_{g}, \beta_{c}\}$ have been empirically determine for each dataset, with initial parameters inspired from \citet{park_few-shot_2019}. These loss functions are explain in the following paragraphs.

\paragraph{Image Reconstruction.}

We train the encoder-decoder using a standard pixel-wise reconstruction loss that minimizes the mean squared error (MSE) between the input image $I$ and its reconstruction $\hat{I}$:

\begin{equation}
    \mathcal{L}_{\text{r}}(I, \hat{I}) = \frac{1}{|I|} \sum_{u \in I, \hat{u} \in \hat{I}} (u - \hat{u})^2,
\end{equation} %
where $u$ and $\hat{u}$ represent individual pixel values in the original and reconstructed images, respectively. This training objective encourages the encoder to learn a compact latent space that captures the essential visual structures of the eye region, including features relevant to gaze direction.

\paragraph{Weighted L2 Loss for 2D PoG Estimation.}

Given ground truth gaze positions \( \mathbf{g}_{\text{true}} \) and predicted values \( \mathbf{g}_{\text{pred}} \), we compute the weighted L2 loss as:

\begin{equation}
\mathcal{L}_{\text{g}} = 
\frac{1}{B} \sum_{i=1}^{B} w_i \left\| \mathbf{g}_{\text{pred}}^{(i)} - \mathbf{g}_{\text{true}}^{(i)} \right\|^2,
\end{equation} %
where \( B \) is the batch size and \( w_i \) is the scalar weight for sample \( i \). The weight is obtained from a precomputed $30 \times 30$ grid of inverse frequency for each dataset to address ground truth region imbalance. Additional details about sample weights are provided in the supplementary materials.

\paragraph{Embedding Consistency Loss.}

To encourage gaze-consistent embeddings, we define a contrastive-style loss that aligns pairwise distances in embedding space with pairwise distances in normalized 2D gaze space.

We first compute the normalized Euclidean distance $\delta_{ij}$ between the PoG gaze $\textbf{g}$:
\begin{equation}
    \delta_{ij} = \frac{\left\| \mathbf{g}_i - \mathbf{g}_j \right\|_2}{\max_{i,j} \left\| \mathbf{g}_i - \mathbf{g}_j \right\|_2 + \varepsilon}
\end{equation}
where \( \varepsilon \) is a small constant for numerical stability. 

Then, the loss is defined as:
\begin{equation}
\mathcal{L}_{\text{c}} = 
\frac{1}{B^2} \sum_{i=1}^B \sum_{j=1}^B w_{ij} 
\left( 
\left\| \mathbf{z}_i - \mathbf{z}_j \right\|_2 - \delta_{ij}
\right)^2,
\end{equation}%
where \( z_i \), \( z_j \) are corresponding latent embeddings. 
This loss penalizes discrepancies between the relative distances in embedding space and the normalized distances in PoG space. The goal is to structure the embedding space such that its geometry reflects gaze similarity, thereby improving downstream generalization and calibration. Additional discussion on the effectiveness of gaze-based loss functions is provided in the supplementary material. 

\subsubsection{Adaptable Gaze Estimator}

To enable rapid personalization with minimal calibration, we adopt a MAML framework to train a gaze estimator \( \mathcal{M}_\theta \) that quickly adapts to new users. The goal is to learn meta-parameters \( \theta^\ast \) such that, with only a few gradient steps on a small calibration set, the model performs well on unseen users.

We frame gaze estimation as a distribution over tasks, each representing gaze prediction for an individual. Each task comprises a small support set (\( k \leq 9 \)) and a corresponding query set from the same participant. Training minimizes post-adaptation validation loss, encouraging generalization to new users rather than overfitting to training data.

Let \( \mathcal{S}_{\text{train}} \) and \( \mathcal{S}_{\text{test}} \) denote disjoint sets of users for meta-training and meta-testing. In each training iteration, a user \( P_i \in \mathcal{S}_{\text{train}} \) is sampled, and a task \( \mathcal{T}_i = \{\mathcal{D}_i^{\text{s}}, \mathcal{D}_i^{\text{q}}\} \) is constructed. The support set \( \mathcal{D}_i^{\text{s}} = \{(\mathbf{z}_j, \mathbf{h}_j, \mathbf{g}_j, w_j)\}_{j=1}^k \) includes latent embeddings, head poses \( \mathbf{h}_j = [R_j \mid t_j] \), 2D gaze targets, and sample weights. The query set \( \mathcal{D}_i^{\text{q}} \) contains \( l \) additional samples from the same user.

From initial model parameters \( \theta \), we compute task-adapted weights \( \theta_i' \) via gradient descent on the support set:

\begin{equation}
\theta_i' = \theta - \alpha \nabla_\theta \mathcal{L}_g^i(\theta),
\end{equation} %
where \( \alpha \) is the inner-loop learning rate and \( \mathcal{L}_g^i \) is the weighted L2 gaze loss. The adapted model is then evaluated on \( \mathcal{D}_i^{\text{q}} \), producing the meta-loss:

\begin{equation}
\mathcal{L}_{\text{meta}} = \sum_i \mathcal{L}_g^i(\theta_i').
\end{equation}

Meta-gradients update the initialization via:

\begin{equation}
\theta \leftarrow \theta - \eta \nabla_\theta \mathcal{L}_{\text{meta}},
\end{equation} %
with \( \eta \) as the outer-loop learning rate. This process is repeated across tasks to learn the meta-initialization \( \theta^\ast \).

At test time, given a new user \( P_{\text{test}} \in \mathcal{S}_{\text{test}} \) and support set \( \mathcal{D}_{\text{test}}^{\text{s}} \), we adapt \( \theta^\ast \) using:

\begin{equation}
\theta_{\text{test}} = \theta^\ast - \alpha \nabla_\theta \mathcal{L}_g^{\text{test}}(\theta^\ast),
\end{equation} %
yielding a personalized model \( \mathcal{M}_{\theta_{\text{test}}} \) evaluated on a held-out query set. This allows effective adaptation from as few as \( k \leq 9 \) samples, enabling low-effort personalization suitable for real-time, in-browser, and mobile settings.

\section{Implementation}

\subsection{Data Preprocessing}

\begin{table*}[ht]
\centering
\caption{\textbf{Gaze Estimation Benchmark Across Accuracy and Efficiency Metrics.} PoG error (cm) on MPIIFaceGaze ($\mathcal{D}_{\text{M}}$), GazeCapture ($\mathcal{D}_{\text{G}}$), and EyeDiap ($\mathcal{D}_{\text{E}}$), reported alongside GFLOPs, parameter count (M), inference delay (ms), and frame per second (FPS). Lower is better for all metrics except FPS. \textbf{Bold} indicates best performance.}
\begin{tabular}{l|c|c|c|c|c|c|c|c|c}
\toprule
Methods    & Year & Open & $\mathcal{D}_{\text{M}} \downarrow$  & $\mathcal{D}_{\text{G}} \downarrow$   & $\mathcal{D}_{\text{E}} \downarrow$       & GFLOPs $\downarrow$          & \#MParams $\downarrow$      & Delay $\downarrow$         & FPS $\uparrow$             \\
\midrule
Mnist      & 2015 & Yes         & 7.29          & NA            & 9.06          & 0.10         & 1.82         & 3.13          & 319.0           \\
iTracker   & 2016 & Yes         & 7.67          & 2.81          & 10.13         & 3.97          & 6.28         & 33.33         & 30.0            \\
GazeNet    & 2017 & Yes         & 6.62          & NA            & 8.51          & 72.24         & 90.23        & 246.31        & 4.1             \\
FullFace   & 2017 & Yes         & 5.65          & NA            & 7.70          & 29.90         & 190         & 174.55        & 5.7             \\
RT-Gene    & 2018 & Yes         & 5.36          & NA            & 7.19          & 12.21         & 31.66        & 43.48         & 23.0            \\
SAGE       & 2019 & No          & NA            & 2.72          & NA            & \textbf{0.04} & NA            & NA            & NA              \\
TAT        & 2019 & No          & NA            & 2.66          & NA            & NA             & NA            & NA            & NA              \\
DilatedNet & 2019 & Yes         & 5.07          & NA            & 7.36          & 202          & 3.92         & 1207.73       & 0.8             \\
Gaze360    & 2019 & Yes         & 4.66          & NA            & 6.37          & 3.65          & 11.94        & 24.39         & 41.0            \\
CA-Net     & 2020 & No          & 4.90          & NA            & \textbf{6.30} & NA             & NA            & NA            & NA              \\
AFF-Net    & 2021 & Yes         & 4.21          & \textbf{2.30} & 9.25          & 26.14         & 1.94         & 109.81        & 9.1             \\
EFE        & 2023 & No          & \textbf{3.89} & 2.48          & NA            & 17.51             & 120            & 1818.18       & 0.6             \\
BlazeGaze  & 2025 & Yes         & 4.56          & 2.32          & 7.53          & 0.15        & \textbf{0.16} & \textbf{0.88} & \textbf{1137.0} \\
\bottomrule
\end{tabular} \label{tab:performance}
\end{table*}
\newcommand{\gridimg}[1]{%
  \begin{subfigure}{\gridwidth}
    \centering
    \frame{\includegraphics[width=\linewidth]{#1}}%
  \end{subfigure}%
}

\newcommand{\SetGrid}[2]{%
  \def\gridrows{#1}%
  \def\gridcols{#2}%
  \newlength{\gridwidth}%
  \setlength{\gridwidth}{\dimexpr 2\linewidth / #2}

}

\SetGrid{2}{4}   

\begin{figure*}[h]   
  \centering
  \gridimg{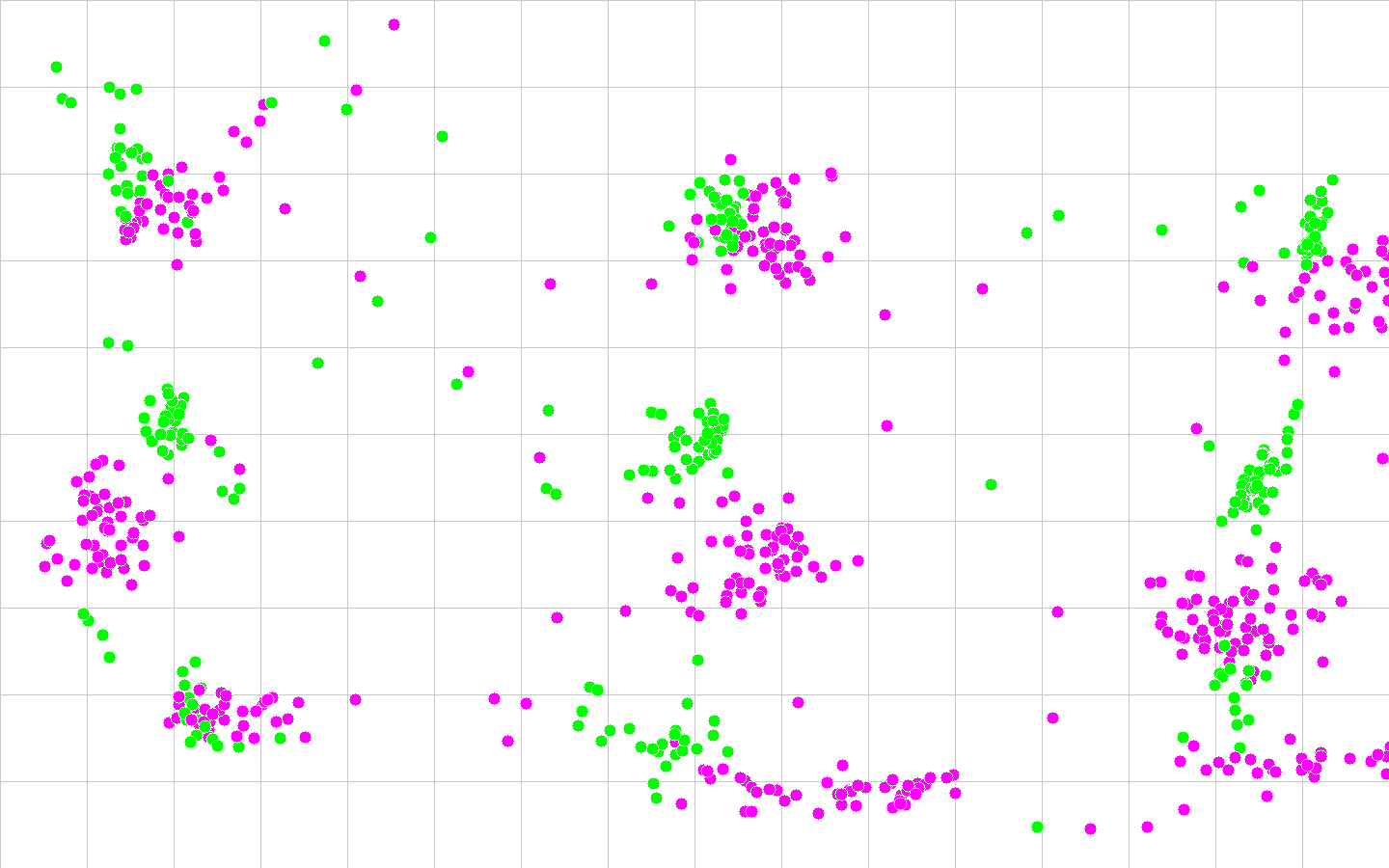}
  \hfill
  \gridimg{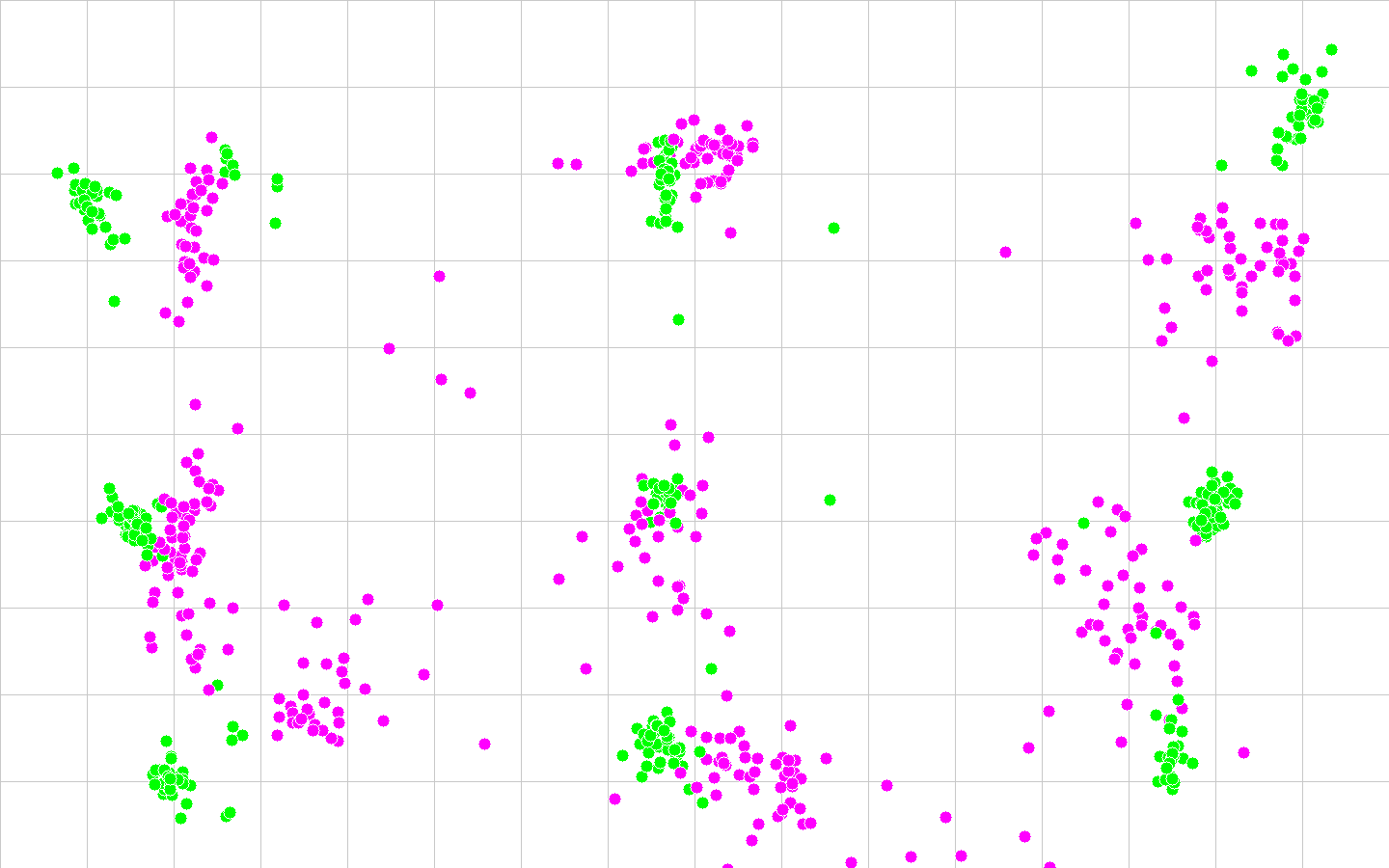}
  \hfill
  \gridimg{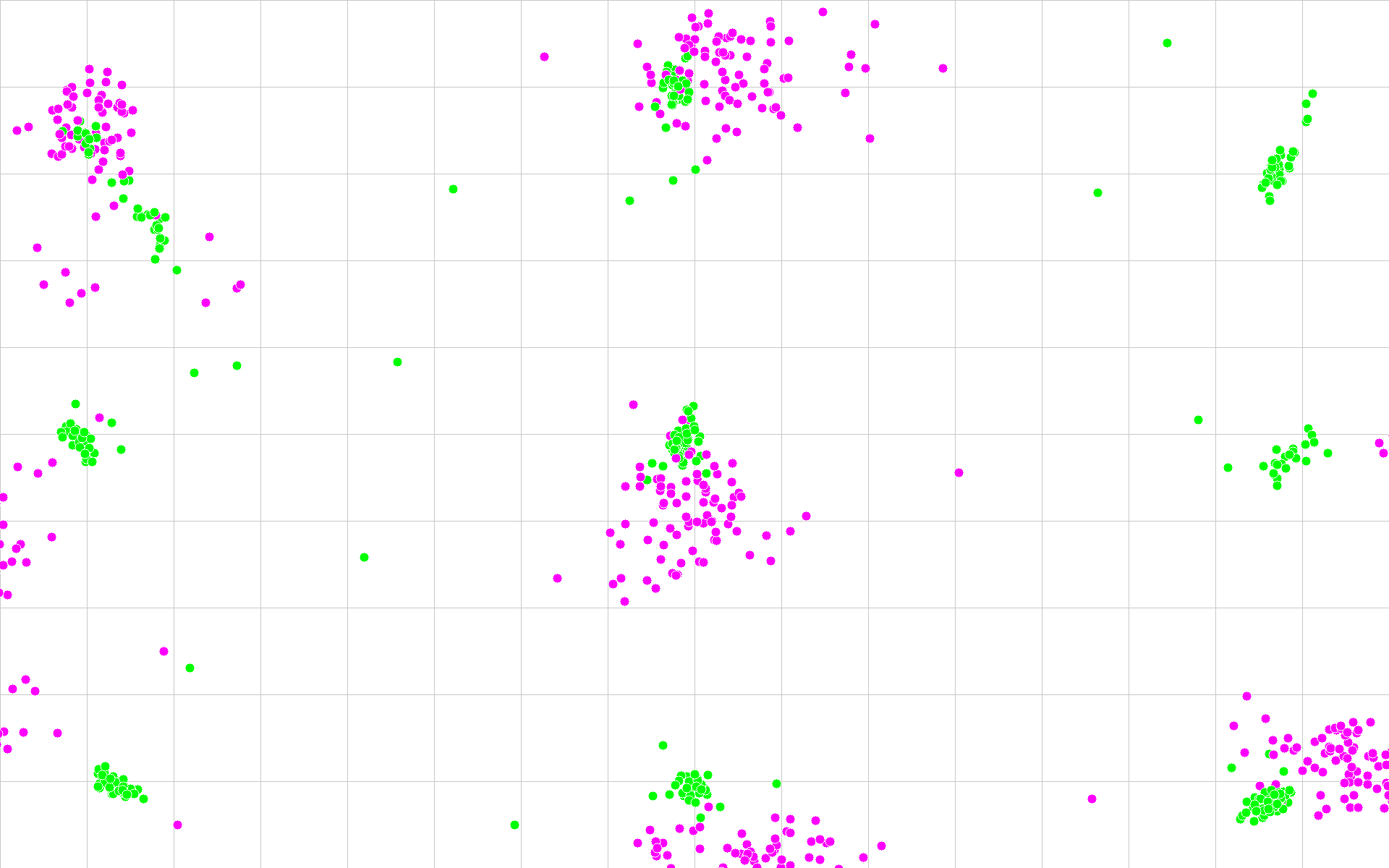}
  \hfill
  \gridimg{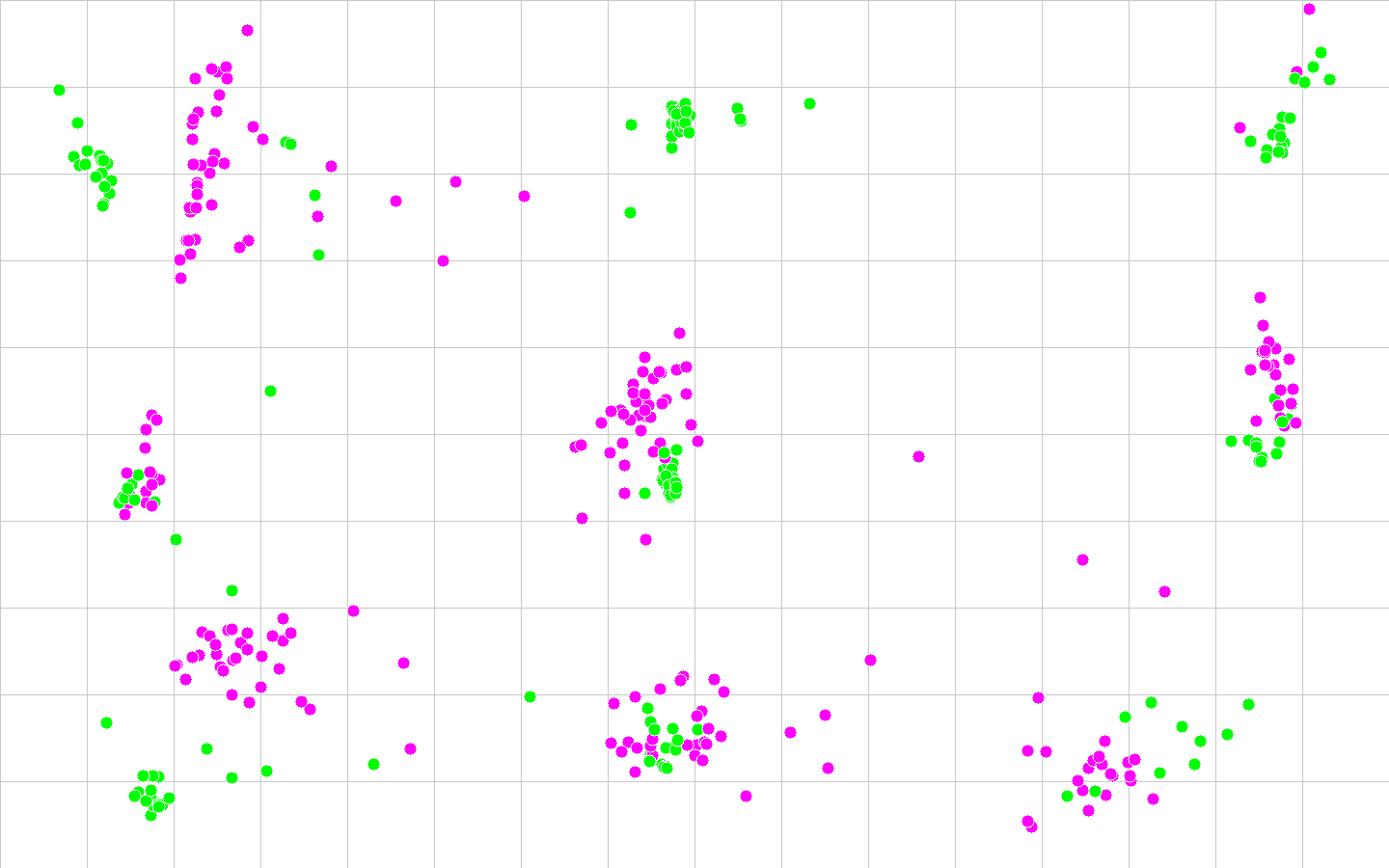}\\[\baselineskip]

  \gridimg{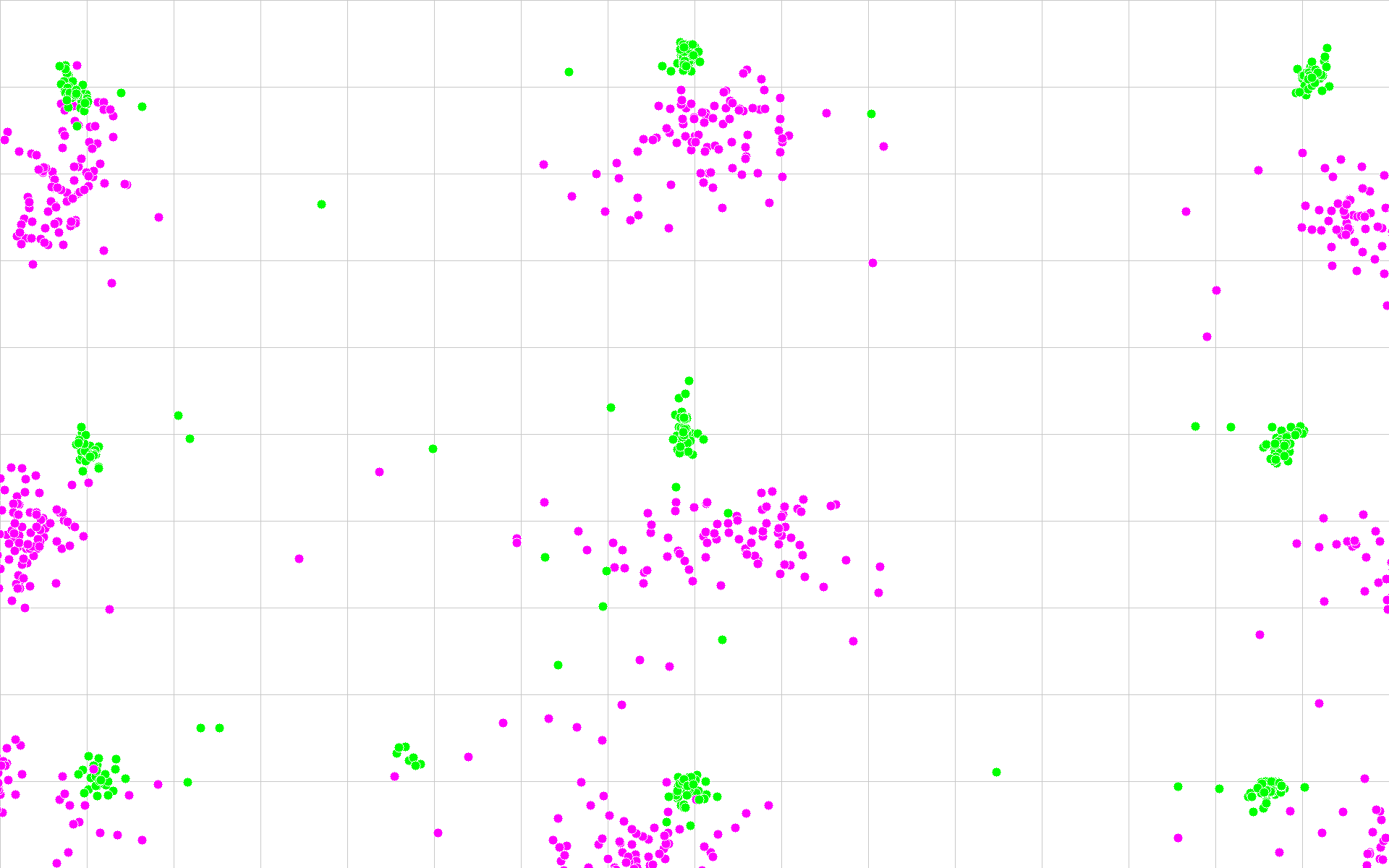}
  \hfill
  \gridimg{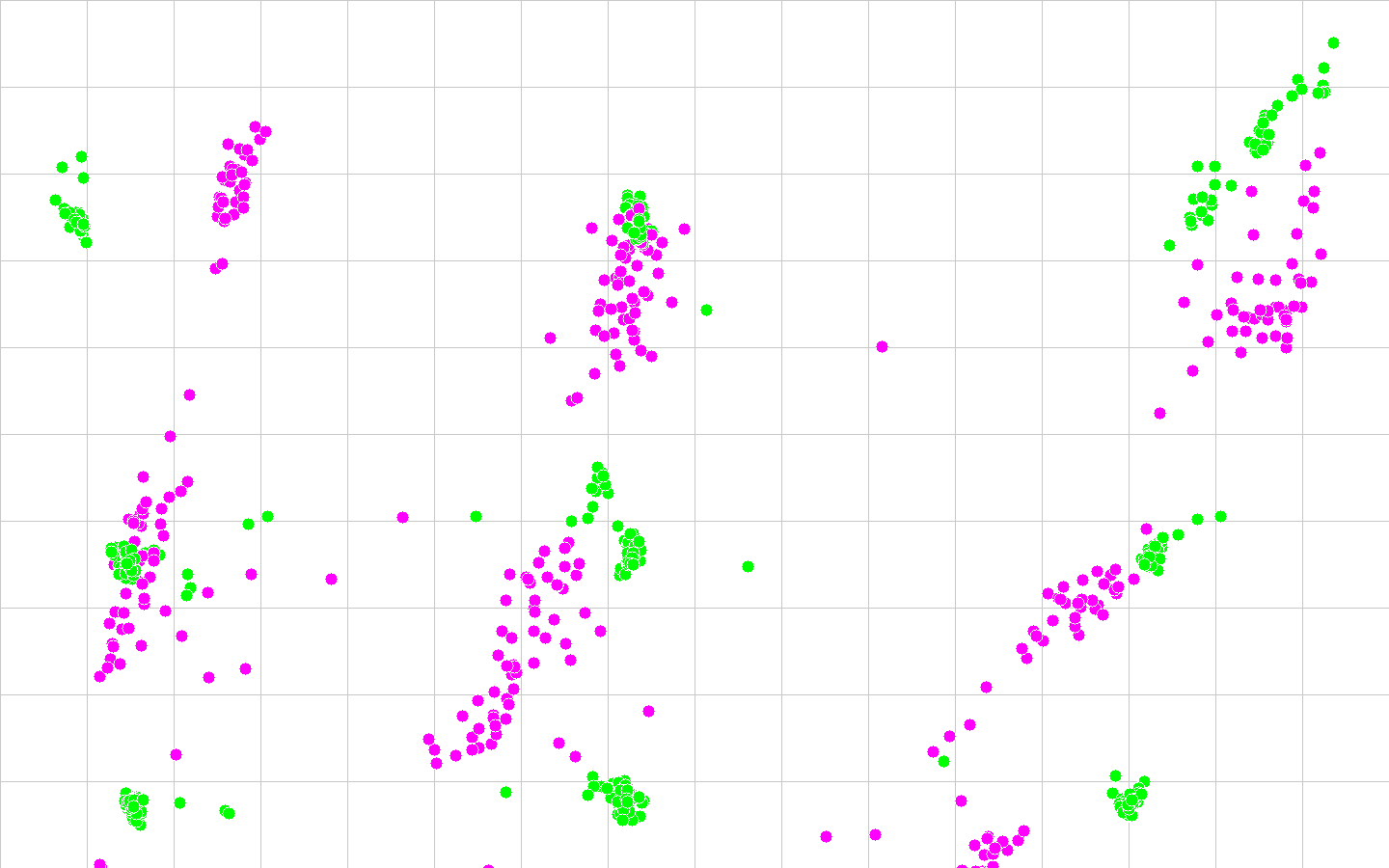}
  \hfill
  \gridimg{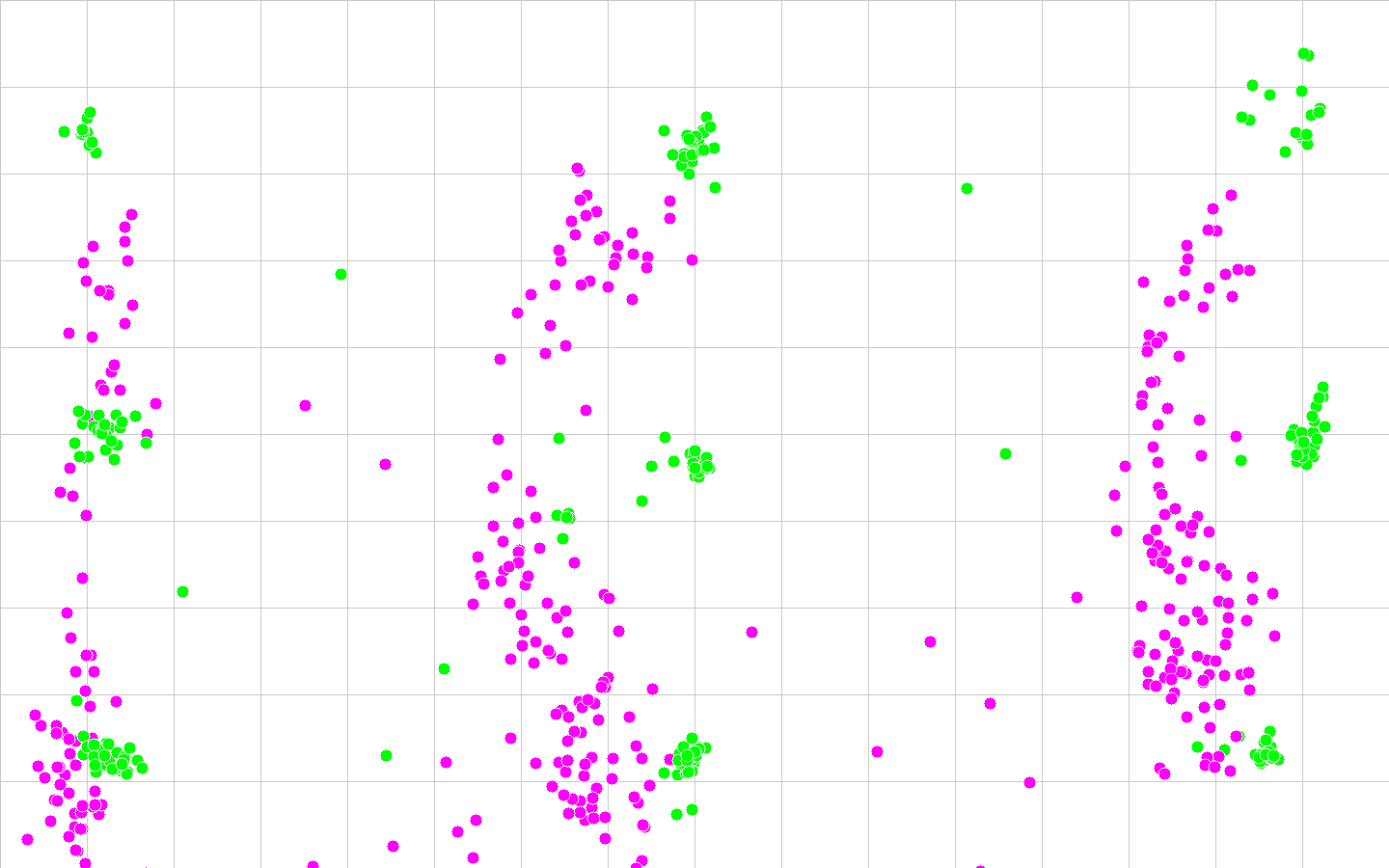}
  \hfill
  \gridimg{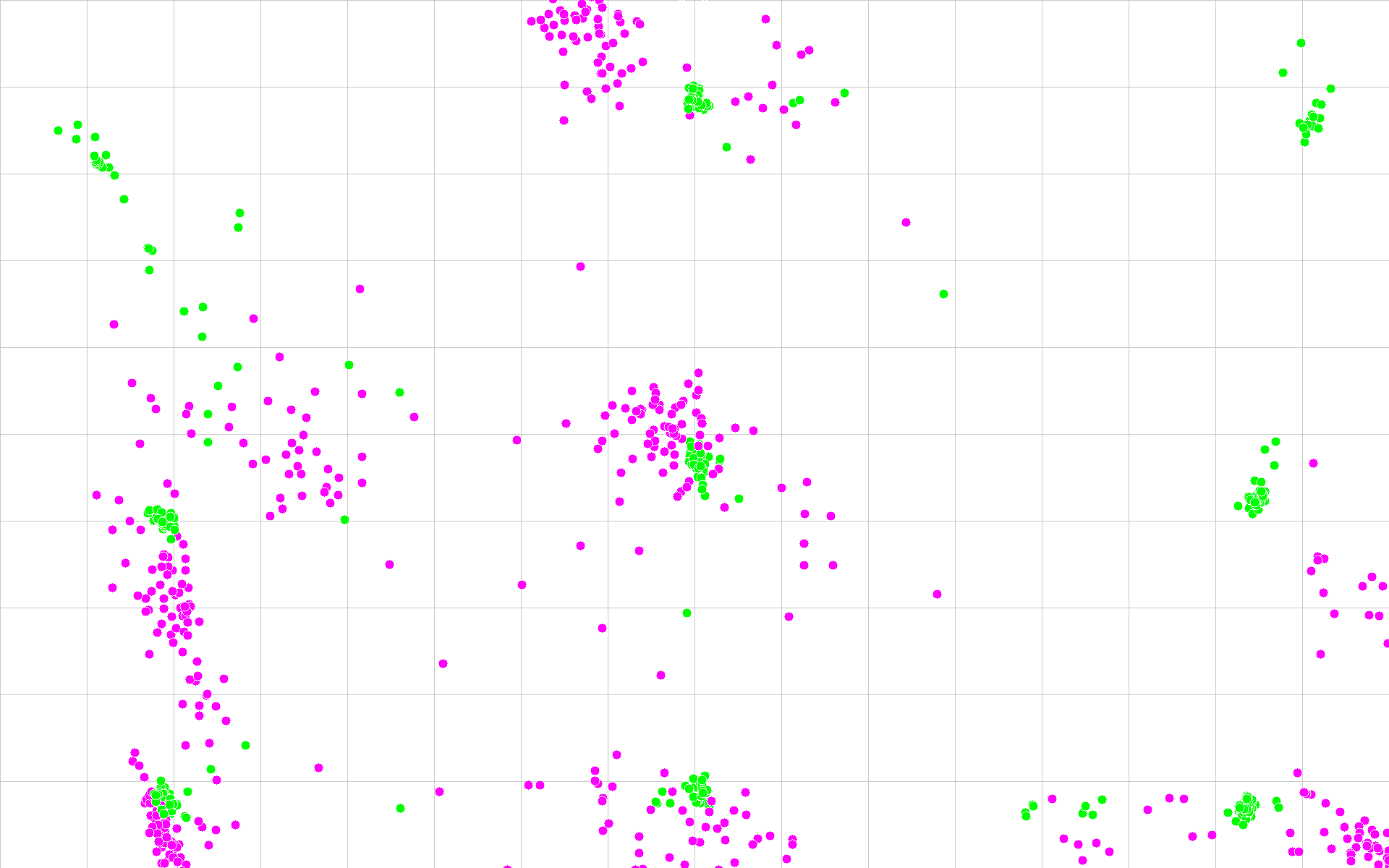}\\[\baselineskip]


  \caption{\textbf{Qualitative Comparison:} Predicted gaze points from \webeyetrack (\textcolor{magenta}{magenta}) vs. ground truth from Tobii X3-120 (\textcolor{green}{green}) during the final Dot Test in the Eye of the Typer dataset.}
  \label{fig:qual_grid}
\end{figure*}
\begin{figure*}[h!]
  \centering

  \begin{subfigure}[b]{0.48\textwidth}
    \centering
    \includegraphics[width=\linewidth]{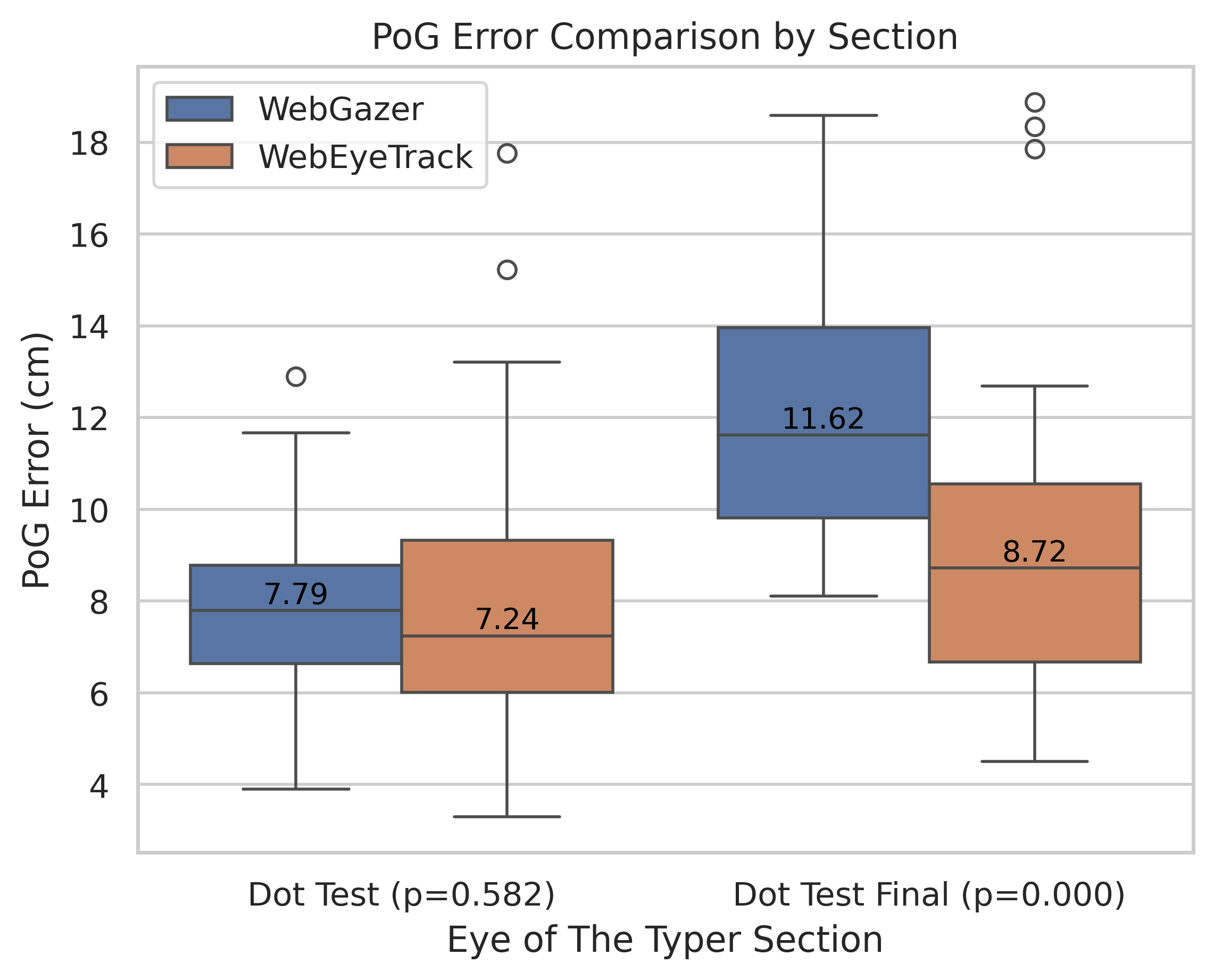}
    \label{fig:subfig:begin_end_accuracy}
  \end{subfigure}
  \hfill
  \begin{subfigure}[b]{0.48\textwidth}
    \centering
    \includegraphics[width=\linewidth]{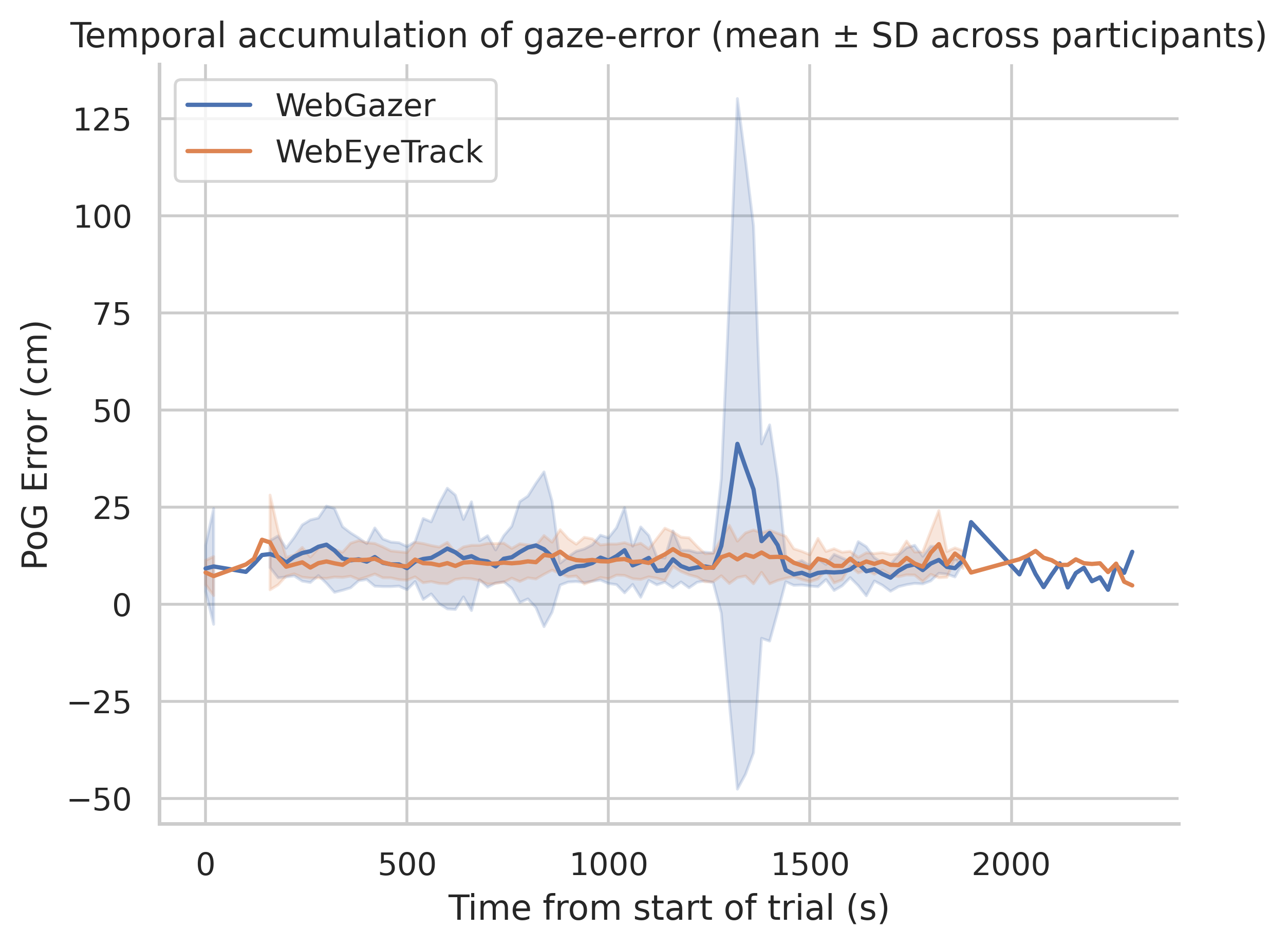}
    \label{fig:subfig:pog_error_over_time}
  \end{subfigure}

  \caption{\textbf{Temporal Accuracy Analysis:} Beginning vs. end accuracy illustrating gaze drift in a 20-minute session (\textbf{left}). Average PoG error over time for \webeyetrack\ and WebGazer, with mean and standard deviation across participants (\textbf{right}).}
  \label{fig:temporal_accuracy_analysis}
\end{figure*}




Our normalization pipeline is tailored for browser-based deployment, where iterative Perspective-n-Point (PnP) solvers and camera intrinsics are unavailable. Instead of using PnP, we compute a homography transformation using the facial landmarks computed by the MediaPipe Facial Landmark Detection model to warp the eye region such that the head appears upright and centered.

\subsection{Neural Networks}

\paragraph{Encoder.} The encoder $\mathcal{E}$ uses a combination of single and double BlazeBlocks for efficient feature extraction from $(128 \times 512 \times 3)$ eye-region images. It outputs a latent embedding of shape $(512)$.

\paragraph{Decoder.} The decoder $\mathcal{D}$ mirrors the encoder structure, using Conv2DTranspose layers to reconstruct the input image.

\paragraph{Gaze Estimator.} The gaze model $\mathcal{M}$ takes the gaze embedding and auxiliary head pose as inputs and propagates their values through a 3-layer multilayer perceptron (MLP) of sizes 16, 16, and 2.

\subsection{Training}

\paragraph{Representation Learning.} BlazeGaze is first trained to map inputs to a gaze-aware embedding space. Training runs for 20 epochs with a batch size of 8, using Adam optimizer and exponential learning rate decay $1 \times 10^{-3}$ and a decay rate of 0.95. The best model is saved, the decoder is discarded, and the encoder is frozen for meta-learning.

\paragraph{Meta-learning.} After learning the representation space, we apply first-order MAML to train $\mathcal{M}$ for rapid personalization. Tasks are defined per user with $k=9$ support and $l=100$ query samples. Inner-loop updates use SGD with $\alpha = 1 \times 10^{-5}$; the outer loop uses Adam with $\eta = 1 \times 10^{-3}$, trained for 1000 steps with 5 inner updates per task.

\paragraph{Deployment.} All components are initially implemented and trained in Python and TensorFlow 2. The trained models are then converted using \verb|tensorflowjs_converter|. The JavaScript version of the framework utilizes Tensorflow.js, specifically the \verb|LayerModels| API\footnote{https://js.tensorflow.org/api/latest/\#class:LayersModel} to support on-device model training and ensure user data never leaves the device. 

\section{Experiments}

We demonstrate the effectiveness and real-world capabilities of \webeyetrack across image and video gaze datasets, illustrating the feasibility of developing SOTA gaze estimation solutions that are compatible with consumer-grade hardware, as our proposed framework achieves competitive accuracy while maintaining minimal computational overhead.

\subsection{Datasets}

We evaluate \webeyetrack on four public datasets covering mobile, webcam, and lab-based scenarios, with consistent preprocessing and normalization for comparability \cite{park_few-shot_2019,balim_efe_2023}.

\paragraph{GazeCapture} Crowdsourced on iOS, $>$2.5M frames from 1450+ users, with screen coordinates normalized via device metadata.

\paragraph{MPIIFaceGaze} Webcam data from 15 subjects ($\sim$3000 frames each) under varied conditions; gaze points normalized to screen dimensions.

\paragraph{EyeDiap} Lab dataset with 16 participants; HD camera footage used with normalized gaze labels.

\paragraph{Eye of the Typer} Tobii Pro X3-120 (120 Hz) eye-tracking during web tasks; we use the “Dot Test” segments with authors’ synchronized preprocessing.






\subsection{Results}

We evaluate \webeyetrack in two settings: (1) within-dataset performance on standard image-based gaze datasets (MPIIFaceGaze, GazeCapture, EyeDiap), and (2) cross-dataset generalization to the Eye of the Typer video dataset. The first setting assesses accuracy and efficiency against SOTA gaze models, while the second compares \webeyetrack\ directly to the leading browser-based solution, WebGazer, under real-world conditions.

\subsubsection{Within-Dataset Accuracy vs. Inference Time}

We perform two-stage training on each dataset and report the resulting PoG error and computational performance
in Figure \ref{fig:performance_vs_inference_graph} and Table \ref{tab:performance}. All inference metrics were computed using a 11th Gen Intel (R) Core(TM) i7-11700F @ 2.50Ghz CPU. 


BlazeGaze achieves accuracy comparable to modern, compute-intensive models, while significantly improving runtime efficiency. Notable, it outperforms the previous fastest model (Mnist) with a 256\% increase in FPS. These results highlights a growing trend in the field: recent models tend to prioritize large architectures (e.g., CNNs, ViTs), often sacrificing real-time viability. In contrast, BlazeGaze offers an effective trade-off between accuracy and speed (see Fig. \ref{fig:performance_vs_inference_graph}), making it suitable for real-world deployment. Additional runtime benchmarks on laptops and mobile devices are provided in the supplementary material.

\subsection{Cross-Dataset Evaluation: Eye of the Typer}

To test generalization, we evaluate BlazeGaze trained on MPIIFaceGaze against the Eye of the Typer dataset to measure the model's robustness to temporal and unseen person gaze estimation. The dataset includes five activities per session: $3\times3$ Dot Tests at the beginning and end for calibration and evaluation, and intermediate tasks involving typing and reading. \webeyetrack is calibrated per participant using only the initial Dot Test (9-point grid).

\subsubsection{Qualitative Gaze Visualizations}

While most gaze benchmarks rely on single-frame image datasets, temporal consistency is rarely assessed. In Fig.~\ref{fig:qual_grid}, we visualize gaze predictions from the final Dot Test. Despite 20-minute sessions, \webeyetrack's predicted gaze points remain closely aligned with ground truth from the Tobii eye-tracker, demonstrating strong spatial and temporal consistency.

\subsubsection{Temporal Accuracy and Degradation}

Fig.~\ref{fig:temporal_accuracy_analysis} shows the change in L1 PoG error between the initial and final Dot Tests. WebGazer's error rises 49\% (7.79 cm$\rightarrow$ 11.62 cm), while \webeyetrack\ increases only 20\% (7.24 cm $\rightarrow$ 8.72 cm). A Mann–Whitney $U$ test confirms that WebEyeTrack’s final error is significantly lower ($p<0.05$), indicating better robustness to temporal drift.

WebGazer’s larger drop stems from its reliance on click-based recalibration, which lacks head-pose modeling and fails during uninterrupted tasks like typing. As shown in Fig.~\ref{fig:temporal_accuracy_analysis}, PoG error over time (10s windows) reveals WebGazer's mid-session instability, while \webeyetrack maintains lower variance and greater temporal stability.

\section{Conclusion}

We presented \webeyetrack, a real-time, browser-compatible gaze estimation framework that integrates metric head pose estimation with few-shot personalization. By combining lightweight representation learning, meta-learning, and efficient in-browser execution, \webeyetrack achieves competitive accuracy with low computational overhead, making it suitable for deployment on consumer devices. Evaluations across multiple datasets demonstrate robust performance across users and time. Future work includes improving model fairness and reducing energy consumption.

\ifanon
\else
    \section{Acknowledgments}
    The research reported here was supported by the Institute of Education Sciences, U.S. Department of Education, through Grant R305A150199 and R305A210347 to Vanderbilt University. The opinions expressed are those of the authors and do not represent views of the Institute or the U.S. Department of Education.
\fi

\bibliography{zotero_webeyetrack_references,manual}

\end{document}